\documentclass[letterpaper]{article} %DO NOT CHANGE THIS
\usepackage{iclr2019_conference,times}
\usepackage{amsmath,amsfonts,bm}

\def\eqref#1{equation~\ref{#1}}
\def\1{\bm{1}}

\DeclareMathAlphabet{\mathsfit}{\encodingdefault}{\sfdefault}{m}{sl}
\SetMathAlphabet{\mathsfit}{bold}{\encodingdefault}{\sfdefault}{bx}{n}

\DeclareMathOperator*{\argmax}{arg\,max}

\usepackage{helvet}  %Required
\usepackage{courier}  %Required
\usepackage{url}  %Required
\usepackage{graphicx}  %Required
\usepackage{booktabs}
\usepackage{microtype}

\usepackage{subcaption}
\usepackage{courier}
\usepackage{amsthm}
\usepackage{color}
\usepackage{listings}
\usepackage{epstopdf}
\usepackage{epsfig}
\usepackage{amsmath}
\usepackage{multirow}
\usepackage{adjustbox}
\usepackage{enumitem}
\usepackage{amssymb}
\usepackage{caption}

\newtheorem*{verblist*}{Verb List}

\newtheorem{example}{Example}

\newcommand{\nop}[1]{}
\newcommand{\cwy}[1]{{#1}}
\newcommand{\zgy}[1]{{#1}}

\begin{document}

\nop{
\twocolumn[
\icmltitle{Recurrent Transfer Learning for Recurrent Neural Networks}
\icmlsetsymbol{equal}{*}

\begin{icmlauthorlist}
\icmlauthor{Aeiau Zzzz}{equal,to}
\icmlauthor{Bauiu C.~Yyyy}{equal,to,goo}
\icmlauthor{Cieua Vvvvv}{goo}
\icmlauthor{Iaesut Saoeu}{ed}
\icmlauthor{Fiuea Rrrr}{to}
\icmlauthor{Tateu H.~Yasehe}{ed,to,goo}
\icmlauthor{Aaoeu Iasoh}{goo}
\icmlauthor{Buiui Eueu}{ed}
\icmlauthor{Aeuia Zzzz}{ed}
\icmlauthor{Bieea C.~Yyyy}{to,goo}
\icmlauthor{Teoau Xxxx}{ed}
\icmlauthor{Eee Pppp}{ed}
\end{icmlauthorlist}

\icmlaffiliation{to}{Department of Computation, University of Torontoland, Torontoland, Canada}
\icmlaffiliation{goo}{Googol ShallowMind, New London, Michigan, USA}
\icmlaffiliation{ed}{School of Computation, University of Edenborrow, Edenborrow, United Kingdom}

\icmlcorrespondingauthor{Cieua Vvvvv}{c.vvvvv@googol.com}
\icmlcorrespondingauthor{Eee Pppp}{ep@eden.co.uk}

\icmlkeywords{Machine Learning, ICML}

\vskip 0.3in
]
}

\iclrfinalcopy

\title{Transfer Learning for Sequences via Learning to Collocate}

\author{
\centerline{Wanyun Cui$^{\S}$ \; Guangyu Zheng$^{\ddag}$ \;   Zhiqiang Shen$^{\P}$ \; Sihang Jiang$^{\ddag}$\; Wei Wang$^{\ddag}$ \;}\\
\centerline{cui.wanyun@sufe.edu.cn,
\{simonzheng96, zhiqiangshen0214, tedjiangfdu\}@gmail.com}\\
\centerline{weiwang1@fudan.edu.cn}\\
{$^{\S}$Shanghai University of Finance and Economics} \\
{$^{\ddag}$Shanghai Key Laboratory of Data Science, Fudan University} \\
{$^{\P}$Shanghai Key Laboratory of Intelligent Information Processing, Fudan University}
}

\maketitle

\begin{abstract}

{\cwy Transfer learning aims to solve the data sparsity for a target domain by applying information of the source domain. Given a sequence (e.g. a natural language sentence), the transfer learning, usually enabled by recurrent neural network (RNN), represents the sequential information transfer. RNN uses a chain of repeating cells to model the sequence data. However, previous studies of neural network based transfer learning simply represents the whole sentence by a single vector, which is unfeasible for seq2seq and sequence labeling. Meanwhile, such layer-wise transfer learning mechanisms lose the fine-grained cell-level information from the source domain.

In this paper, we proposed the \underline{a}ligned \underline{r}ecurrent \underline{t}ransfer, ART, to achieve cell-level information transfer. ART is under the pre-training framework. Each cell attentively accepts transferred information from a set of positions in the source domain. Therefore, ART learns the cross-domain word collocations in a more flexible way. We conducted extensive experiments on both sequence labeling tasks (POS tagging, NER) and sentence classification (sentiment analysis). ART outperforms the state-of-the-arts over all experiments.}

% transfer information among modules in the same RNN layer. An RNN layer models the sequence data through sequential memory. Similarly, we build the transfer mechanism for sequential memory transfer between each two adjacent units from different domains. We propose the transferable RNN, a novel mechanism for sequential memory transfer in RNN. It uses two extra connections to attentively transfer the sequence memory from the source domain to the target domain. We implement the transfer mechanism over LSTM and GRU. Both implementations beat the non-recurrent transfer over different NLP tasks.
\end{abstract}

% !TEX root = transfernlp.tex

\section{Introduction}

Most previous NLP studies focus on open domain tasks. But due to the variety and ambiguity of natural language~\citep{glorot2011domain,song2011short}, models for one domain usually incur more errors when adapting to another domain. This is even worse for neural networks since embedding-based neural network models usually suffer from overfitting~\citep{peng-EtAl}. While existing NLP models are usually trained by the open domain, they suffer from severe performance degeneration when adapting to specific domains. This motivates us to train specific models for specific domains.

The key issue of training a specific domain is the insufficiency of labeled data. Transfer learning is one promising way to solve the insufficiency~\citep{jiang2007instance}. Existing studies~\citep{daume2009frustratingly,jiang2007instance} have shown that (1) NLP models in different domains still share many common features (e.g. common vocabularies, similar word semantics, similar sentence syntaxes), and (2) the corpus of the open domain is usually much richer than that of a specific domain.  %So one promising way is transfer learning from the open domain to specific domains. %~\citep{pan2010survey}.

%The recurrent neural network (RNN) is a widely used neural network in representing a natural language sentence. It uses an identical recurrent weight matrix across all time stamps (words). Each RNN cell accepts the information from the previous time stamp. Precisely capturing and passing the information for each cell is the key for RNN.

Our transfer learning model is under the pre-training framework. We first pre-train the model for the source domain. Then we fine-tune the model for the target domain. Recently, some pre-trained models (e.g. BERT~\citep{devlin2018bert}, ELMo~\citep{peters2018deep}, GPT-2~\citep{radford2019language}) successfully learns general knowledge for text. The difference is that these models use a large scale and domain-independent corpus for pre-training. In this paper, we use a small scale but domain-dependent corpus as the source domain for pre-training. We argue that, for the pre-training corpus, the domain relevance will overcome the disadvantage of limited scale.

% Unlike their problem settings, in transfer learning, we focus on limited labeled corpus of the source domain. we use the pre-trained information from labeled data from the same task but with different domains. Although the available training samples is much smaller than these unlabeled data, we argue that the pre-trained model still helps because their task are quite close.

\begin{figure}[ht]
\centering
\includegraphics[scale=0.7]{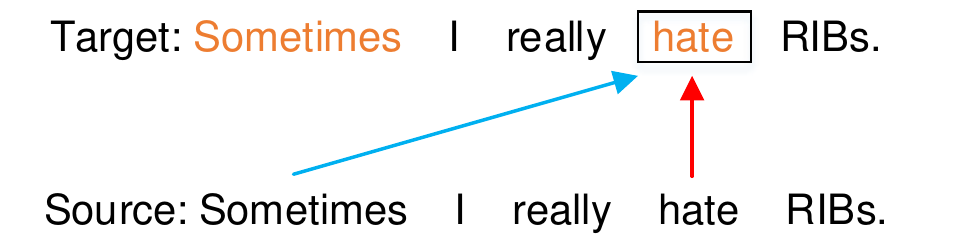}
\caption{Motivation of ART. The orange words ``sometimes'' and ``hate'' are with insufficiently trained parameters. The red line indicates the information transfer from the corresponding position. The blue line indicates the information transfer from a collocated word.}
\label{fig:intuition} %% label for entire figure
\vspace{-0.5cm}
\end{figure}

Most previous transfer learning approaches~\citep{li2018hierarchical,ganin2016domain} only transfer information across the whole layers. This causes the information loss from cells in the source domain. {\cwy ``Layer-wise transfer learning'' indicates that the approach represents the whole sentence by a single vector. So the transfer mechanism is only applied to the vector.} We highlight the effectiveness of {\it precisely capturing and transferring information of each cell from the source domain} in two cases. First, in seq2seq (e.g. machine translation) or sequence labeling (e.g. POS tagging) tasks, all cells directly affect the results. So layer-wise information transfer is unfeasible for these tasks. Second, even for the sentence classification, cells in the source domain provide more fine-grained information to understand the target domain. For example, in figure~\ref{fig:intuition}, parameters for ``hate'' are insufficiently trained. The model transfers the state of ``hate'' from the source domain to understand it better.

Besides transferring the corresponding position's information, the transfer learning algorithm captures the {\bf cross-domain long-term dependency}. Two words that have a strong dependency on each other can have a long gap between them. Being in the insufficiently trained target domain, a word needs to represent its precise meaning by incorporating the information from its collocated words. {\cwy Here ``collocate'' indicates that a word's semantics can have long-term dependency on other words. To understand a word in the target domain, we need to precisely represent its collocated words from the source domain. We learn the collocated words via the attention mechanism~\citep{bahdanau2014neural}.} For example, in figure~\ref{fig:intuition}, ``hate'' is modified by the adverb ``sometimes'', which implies the act of hating is not serious. But the ``sometimes'' in the target domain is trained insufficiently. We need to transfer the semantics of ``sometimes'' in the source domain to understand the implication. Therefore, we need to carefully align word collocations between the source domain and the target domain to represent the long-term dependency. %And we need to carefully determine what information to transfer between the aligned cells.

In this paper, we proposed ART (\underline{a}ligned \underline{r}ecurrent \underline{t}ransfer), a novel transfer learning mechanism, to transfer cell-level information by learning to collocate cross-domain words. ART allows the {\bf cell-level information transfer} by directly extending each RNN cell. ART incorporates the hidden state representation corresponding to the same position and a function of the hidden states for all words weighted by their attention scores.
%It discriminates information transfer from the corresponding position and all positions with collocated words.

{\bf Cell-Level Recurrent Transfer} ART extends each recurrent cell by taking the states from the source domain as an extra input. %In the original RNN, each cell receives two inputs: one from the lower layer and one from the previous time stamp.
While traditional layer-wise transfer learning approaches discard states of the intermediate cells,
%In traditional layer-wise transfer learning, the states of the intermediate cells are discarded.
{\cwy ART uses cell-level information transfer, which means each cell is affected by the transferred information. For example, in figure~\ref{fig:intuition}, the state of ``hate'' in the target domain is affected by ``sometimes'' and ``hate'' in the source domain.} Thus ART transfers more fine-grained information.

%the previous cell in the target domain still pass information to the next cell. And the information from the source domain also transfers to the target domain. ART benefits from the cell-level transferred information.

\nop{
sequential memory transferring through all cells, we modify the recurrent layer. In the original RNN, each cell accepts two inputs: one from the lower layer and one from the previous time stamp. The data from the previous time stamp is called the {\it sequential memory}. We extend the recurrent cell by adding the sequential memory from the source domain as the extra input.
Besides, the transferable recurrent unit also accepts inputs from the source domain's lower layer (usually an embedding layer) and the previous time stamp.
The source domain's neural network still uses the original recurrent cells, which generates standard sequential memory. The target domain uses the modified RNN cells above to allow cell-level information transfer. Similar to the RNN mechanism, each modified RNN cell has the same transfer mechanism and weight matrices. We highlight that, the modified recurrent cells benefit from the transferred sequential memory.
}

{\bf Learn to Collocate and Transfer} For each word in the target domain, ART learns to incorporate two types of information from the source domain: (a) the hidden state corresponding to the same word, and (b) the hidden states for all words in the sequence.
%Information (a) represents the semantics of the same word in the source domain.
Information (b) enables ART to capture the cross-domain long-term dependency. ART learns to incorporate information (b) based on the attention scores~\citep{bahdanau2014neural} of all words from the source domain.
%\nop{\bf For example, in figure~\ref{fig:intuition}, for word ``hate'' of the target domain, the red line transfers its corresponding semantics of the source domain. The blue line transfers a collocated word ``sometimes'' of the source domain. So it represent the cross domain dependency between ``hate'' and ``sometimes''.} %we compute the correlation for each word pair.
%We sum up the information from all cells in the source domain according to their correlations.
%Each RNN cell has a position, which indicates its order in the sequence. Each cell in the target domain attentively selects a set of positions in the source domain. ART collocates words of the positions and transfers their information.
Before learning to transfer, we first pre-train the neural network of the source domain. Therefore ART is able to leverage the pre-trained information from the source domain. %Through the pre-training, its states represent the semantics of the source domain. Therefore, ART successfully represents the cross-domain long-term dependency.

\section{Aligned Recurrent Transfer}

%We use LSTM networks based on PNN network framework, to solve the transfer learning problem of sentence understanding. The input is a collection of sentences from open domain and target domain. The output can be a single label (e.g. sentiment analysis of the sentence) or a label sequence of the question (e.g. POS tagging).

In this section, we elaborate the general architecture of ART. We will show that, ART precisely learns to collocate words from the source domain and to transfer their cell-level information for the target domain.

{\bf Architecture }
The source domain and the target domain share an RNN layer, from which the common information is transferred. We pre-train the neural network of the source domain. Therefore the shared RNN layer represents the semantics of the source domain. The target domain has an additional RNN layer. Each cell in it accepts transferred information through the shared RNN layer. Such information consists of (1) the information of the same word in the source domain (the red edge in figure 2); and (2) the information of all its collocated words (the blue edges in figure 2). ART uses attention~\citep{bahdanau2014neural} to decide the weights of all candidate collocations. The RNN cell controls the weights between (1) and (2) by an update gate.

%Both the source domain and the target domain use RNN to represent the sequence. In the RNN layer of the target domain, each cell accepts extra transferred information from the source domain. The transferred information for the $i$-the cell consists of (1) the information from the $i$-th cell in the source domain (the red edge in figure~\ref{fig:modelillustration}), which  and (2) the information from all collocated positions (the blue edges in figure~\ref{fig:modelillustration}). The RNN cell uses an update gate to control the weights between (1) and (2). For (2), the cell uses the attention mechanism~\citep{bahdanau2014neural} to collocate with a set of positions in the source domain and transfer their information.}

\begin{figure}[ht]
\centering
\includegraphics[scale=0.6]{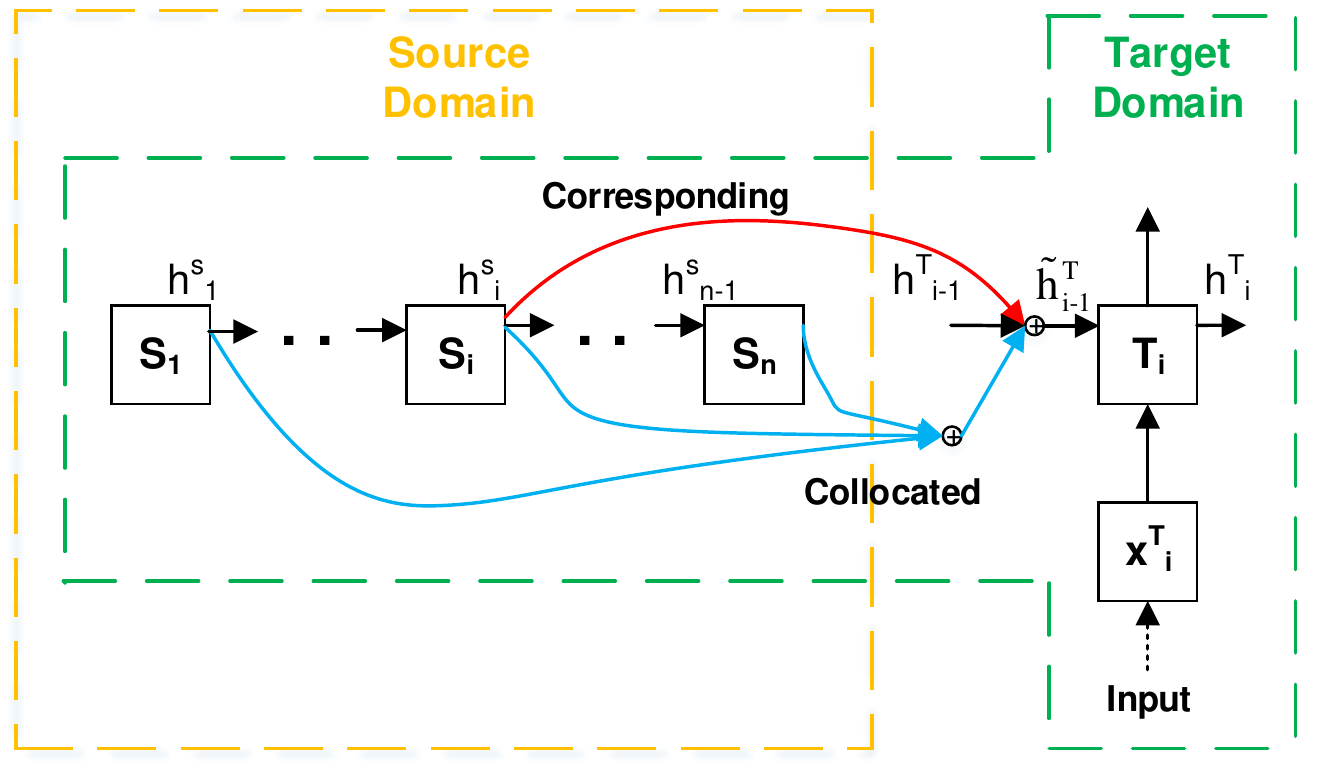}
\caption{ART architecture.}
\label{fig:modelillustration} %% label for entire figure
\vspace{-0.2cm}
\end{figure}

Figure~\ref{fig:modelillustration} shows the architecture of ART. The yellow box contains the neural network for the source domain, which is a classical RNN. The green box contains the neural network for the target domain. $S_i$ and $T_i$ are cells for the source domain and target domain, respectively. $T_i$ takes $x_i$ as the input, which is usually a word embedding. The two neural networks overlap each other. The source domain's neural network transfers information through the overlapping modules. We will describe the details of the architecture below. Note that although ART is only deployed over RNNs in this paper, its attentive transfer mechanism is easy to be deploy over other structures (e.g. Transformer~\citep{vaswani2017attention})

%that the yellow box and the green box have some overlappings. So the target domain's neural network uses the recurrent layer in the source domain. The hidden state in the source domain is treated as the shared knowledge and passed to the target domain. We will describe the details of the architecture below.

{\bf RNN for the Source Domain} The neural network of the source domain consists of a standard RNN. Each cell $S_i$ captures information from the previous time step $h^S_{i-1}$, computes and passes the information $h^S_i$ to the next time step. More formally, we define the performance of $S_i$ as:
\begin{equation}
\small
h_i^S=RNN(h_{i-1}^S,x^S_i;\theta_S)
\end{equation}
%where $h_i^S$ is the hidden state of the $i$-th recurrent cell,
where $\theta_S$ is the parameter (recurrent weight matrix).

%Here we say the RNN is static, since its parameters $\theta_S$ are ``freezed'' during training the target domain's model. $\theta_S$ is only updated once by the source domain's training data. This is treated as a pre-processing for training the target domain.

{\bf Information Transfer for the Target Domain} Each RNN cell in the target domain leverages the transferred information from the source domain. Different from the source domain, the $i$-th hidden state in the target domain $h^T_i$ is computed by:
\begin{equation}
\small
h^T_i=RNN(\widetilde{h^T_{i-1}},x^T_i;\theta_T)
\end{equation}
where $\widetilde{h^T_{i-1}}$ contains the information passed from the previous time step in the target domain ($h^T_{i-1}$), and the transferred information from the source domain ($\psi_i$). We compute it by:
\begin{equation}
\small
\widetilde{h^T_{i-1}}=f(h^T_{i-1}, \psi_i|\theta_f) % g(h^T_{i-1},h_1^S, \cdots, h_n^S) ) % h_1^S, \cdots, h_n^S)
\end{equation}
where $\theta_f$ is the parameter for $f$. %We will elaborate them later. %$h^T_{i-1}$ in $g()$ is used to assign different attentions for different source domain's positions.

Note that both domains use the same RNN function with different parameters ($\theta_S$ and $\theta_T$). Intuitively, we always want to transfer the common information across domains. And we think it's easier to represent common and shareable information with an identical network structure.

{\bf Learn to Collocate and Transfer} We compute $\psi_i$ by aligning its collocations in the source domain. We consider two kinds of alignments: (1) The alignment from the corresponding position. This makes sense since the corresponding position has the corresponding information of the source domain. (2) The alignments from all collocated words of the source domain. This alignment is used to represent the long-term dependency across domains. We use a ``concentrate gate'' $u_i$ to control the ratio between the corresponding position and collocated positions. We compute $\psi_i$ by:
\begin{equation}
\small
\psi_i= (1-u_i) \circ \pi_i +  u_i \circ h^S_i
\end{equation}
where
\begin{equation}
\small
u_i = \delta(W_u h^S_i+C_u \pi_i )
\end{equation}
$\pi_i$ denotes the transferred information from collocated words. $\circ$ denotes the element-wise multiplication. $W_u$ and $C_u$ are parameter matrices.

In order to compute $\pi_i$, we use attention~\citep{bahdanau2014neural} to incorporate information of all candidate positions in the sequence from the source domain. We denote the strength of the collocation intensity to position $j$ in the source domain as $\alpha_{ij}$. We merge all information of the source domain by a weighted sum according to the collocation intensity. More specifically, we define $\pi_i$ as:
%Function $g()$ select proper information from $h^T_{i-1}$ and $h_1^S, \cdots, h_n^S$ to transfer. Directly concatenate these vectors will lead to a very high dimension vector. Summing up or averaging will lead to information loss. Inspired by the attention mechanism~\citep{bahdanau2014neural}, we proposed an attentive transfer mechanism. We search for a set of positions in the source domain and the previous cell in the target domain, and concentrated on them. We compute $g(h^T_i,h_1^S, \cdots, h_n^S)$ by:
\begin{equation}
\small
\label{eqn:funcg}
\pi_i=\sum_{j=1}^n\alpha_{ij}h^S_{j}
\end{equation}
where
\begin{equation}
\small
\alpha_{ij} = \frac{\exp(a(h_{i-1}^T,h_j^S))}{\sum_{j'=1}^n\exp(a(h_{i-1}^T,h_{j'}^S))}
\end{equation}
%\begin{equation}
%e_{ij}=a(h_{i-1}^T,h_j^S)
%\end{equation}
$a(h_i^T,h_j^S)$ denotes the collocation intensity between the $i$-th cell in the target domain and the $j$-th cell in the source domain. The model needs to be evaluated $O(n^2)$ times for each sentence, due to the enumeration of $n$ indexes for the source domain and $n$ indexes for the target domain. Here $n$ denotes the sentence length. By following~\cite{bahdanau2014neural}, we use a single-layer perception:
\begin{equation}
\small
a(h_i^T,h_j^S)=v_a^\top\tanh(W_a h_i^T + U_a h^S_j)
\end{equation}
where $W_a$ and $U_a$ are the parameter matrices. %So the $\theta_g$ in Eq~\eqref{eqn:funcg} contains $v_a$, $W_a$ and $U_a$.
Since $U_a h^S_j$ does not depend on $i$, we can pre-compute it to reduce the computation cost.

{\bf Update New State} To compute $f$, we use an update gate $z_i$ to determine how much of the source domain's information $\widetilde{\psi_i}$ should be transferred. $\widetilde{\psi_i}$ is computed by merging the original input $x_i$, the previous cell's hidden state $h^T_{i-1}$ and the transferred information $\psi_{i}$. We use a reset gate $r_i$ to determine how much of $h^T_{i-1}$ should be reset to zero for $\widetilde{\psi_i}$. More specifically, we define $f$ as:
\begin{equation}
\small
f(h^T_i,\psi_i)=(1-z_i)\circ h^T_{i-1} + z_i \circ \widetilde{\psi_i}
\end{equation}
%\begin{equation}
%f(h^T_i,\psi_i)=(1-z_i)\circ h^T_{i-1} + z_i \circ [(1-u_i)\circ h^S_i + u_i \circ \widetilde{\psi_i} ]
%\end{equation}
where
\begin{equation}
\small
\begin{aligned}
\widetilde{\psi_i}=&\tanh (W_\psi x_i+U_\psi[r_i\circ h^T_{i-1}]+C_\psi\psi_i) \\
z_i = &\delta(W_z x_i +U_zh^T_{i-1}+C_z\psi_i) \\
r_i = &\delta(W_r x_i +U_rh^T_{i-1}+C_r\psi_i) \\
\end{aligned}
\end{equation}
\nop{
\begin{equation}
\begin{aligned}
\widetilde{\psi_i}=&\tanh (W_\psi x_i+U_\psi[r_i\circ h^T_{i-1}]+C_\psi[(1-u_i)\circ h^S_i + u_i \circ \psi_i ]) \\
z_i = &\delta(W_z x_i +U_zh^T_{i-1}+C_z[(1-u_i)\circ h^S_i + u_i \circ \psi_i ]) \\
r_i = &\delta(W_r x_i +U_rh^T_{i-1}+C_r[(1-u_i)\circ h^S_i + u_i \circ \psi_i ]) \\
\end{aligned}
\end{equation}
}
Here these $W, U, C$ are parameter matrices.

{\bf Model Training:} {\cwy We first pre-train the parameters of the source domain by its training samples. Then we fine-tune the pre-trained model with additional layers of the target domain. The fine-tuning uses the training samples of the target domain. All parameters are jointly fine-tuned.}

\section{ART over LSTM} %Neural Networks}
\label{sec:lstm}
In this section, we illustrate how we deploy ART over LSTM. LSTM specifically addresses the issue of learning long-term dependency in RNN. Instead of using one hidden state for the sequential memory, each LSTM cell has two hidden states for the long-term and short-term memory. So the ART adaptation needs to separately represent information for the long-term memory and short-term memory.

% design specific ART models. We do this for two models: LSTM and scaled dot-product attention (SDPA)~\cite{vaswani2017attention}. LSTM is a widely used RNN version. SDPA discards CNN or RNN and only uses attention mechanism for sequence representation. We denote the two adapted versions as ART-LSTM and ART-SDPA, respectively.

%{\bf ART-LSTM} Long-short term memory network (LSTM) specifically address the issue of learning long term dependency in RNN. Instead of using one hidden state for the sequential memory, each LSTM cell has two hidden states for the long-term and short-term memory. So the key of the ART adaptation is to retain and transfer information through the two hidden states.

The source domain of ART over LSTM uses a standard LSTM layer. The computation of the $t$-th cell is precisely specified as follows:
\begin{equation}
\small
\label{lstm:1}
\begin{bmatrix}
    \widetilde{c^S_i} \\
    o^S_t \\
    i^S_t \\
    f^S_t
\end{bmatrix}
=
\begin{bmatrix}
    \tanh \\
    \sigma \\
    \sigma \\
    \sigma
\end{bmatrix}
T^S_{A, b}
\begin{bmatrix}
    x^S_t \\
    h^S_{t-1}
\end{bmatrix}
\end{equation}
\begin{equation}
\small
\label{lstm:2}
c^S_t = \widetilde{c^S_t} \circ i^S_t + c^S_{t-1} \circ f^S_t
\end{equation}
\begin{equation}
\small
\label{lstm:3}
h^S_t = o^S_t  \circ \tanh(c^S_t)
\end{equation}
Here $h^S_t$ and $c^S_t$ denote the short-term memory and long-term memory, respectively.
%Each LSTM cell has two hidden state: short-term memory $h^S_t$ and long-term memory $c^S_t$. %We denote the $t$-th short-term memory and long-term memory in the source domain as $h^S_t$ and $c^T_t$, respectively.

In the target domain, we separately incorporate the short-term and long-term memory from the source domain. More formally, we compute the $t$-the cell in the target domain by:
% the attentive knowledge transfer over the short-term memory and long-term memory. This is because we think similar knowledge is easy to transfer. And the same terms' memories work similar. We compute ART-LSTM by:
\begin{equation}
\small
\label{lstm:4}
\begin{bmatrix}
    \widetilde{c^T_i} \\
    o^T_t \\
    i^T_t \\
    f^T_t
\end{bmatrix}
=
\begin{bmatrix}
    \tanh \\
    \sigma \\
    \sigma \\
    \sigma
\end{bmatrix}
T^T_{A, b}
\begin{bmatrix}
    x^T_t \\
    f(h^T_{i-1},\psi_{hi}|\theta_{fh})\\
    %g(h^T_{t-1}, h^S_1, \cdots ,h^S_n;\theta_{LSTM-h}) \\
\end{bmatrix}
\end{equation}
\begin{equation}
\small
c^T_t = \widetilde{c^T_t} \circ i^T_t + f(c^T_{i-1},\psi_{ci}|\theta_{fc}) \circ f^T_t
\end{equation}
\begin{equation}
\small
h^T_t = o^T_t \circ \tanh(c^T_t)
\end{equation}
where $f(h^T_{i-1},\psi_{hi}|\theta_{fh})$ and $f(c^T_{i-1},\psi_{ci}|\theta_{fc})$ are computed by Eq~(\ref{eqn:funcg}) with parameters $\theta_{fh}$ and $\theta_{fc}$, respectively. $\psi_{hi}$ and $\psi_{ci}$ are the transferred information from the short-term memory ($h^S_1 \cdots h^S_n$ in Eq.~(\ref{lstm:2})) and the long-term memory ($c^S_1 \cdots c^S_n$ in Eq.~(\ref{lstm:3})), respectively.

{\bf Bidirectional Network} We use the bidirectional architecture to reach all words' information for each cell. The backward neural network accepts the $x_i (i=1 \dots n)$ in reverse order.
%We denote the $h^T_t$ in equation~\ref{lstm:3} of the forward neural network and backward neural network as $\overrightarrow{h^T_t}$ and $\overleftarrow{h^T_t}$, respectively.
We compute the final output of the ART over LSTM by concatenating the states from the forward neural network and the backward neural network for each cell.

\section{Experiments}
We evaluate our proposed approach over sentence classification (sentiment analysis) and sequence labeling task (POS tagging and NER).

%We use two types of NLP tasks (POS tagging and sentiment analysis) to evaluate our proposed approach. We have mentioned that layer-wise transfer learning cannot capture the transferred information for each cell, and thus is unfeasible for seq2seq or sequence labeling tasks. Thus we first use POS tagging, a typical sequential labeling task, to verify the effectiveness of our approach in sequence labeling problems. We also evaluate our approach in a sentiment analysis problem, which is widely used in cross-domain sentence classification.
%We use the sentiment analysis problem to verify the effectiveness of ART in single-output tasks.
%The results show that (1) {\it our approach successfully uses transferred knowledge from open domain to improve the effectiveness}; (2) {\it our approach performs the best in diverse applications, especially for sequence-to-sequence labeling tasks}.

\subsection{Setup}
%{\cwy machine configuration (CPU, GPU, memory), hyperparameters in the model (number of dimentions, etc.)}

All the experiments run over a computer with Intel Core i7 4.0GHz CPU, 32GB RAM, and a GeForce GTX 1080 Ti GPU.

%{\cwy two baseline}

%{\bf Baselines:} We design three baselines to verify the effectiveness of the proposed approach. Baseline 1 and baseline 2 directly use the source domain's neural network without transfer learning. Baseline1 is trained by the source domain's data. Baseline2 is trained by the target domain's data. We use it to verify if our approach transfers valid knowledge from open domain to improve the results.  We compare with baseline 1 and baseline 2 to verify if our approach beats the source domain's model and the target domain's model, respectively.

%{\bf NLP Applications:} We evaluate our approach over different NLP tasks, including the sentence classification and the sequence-to-sequence labeling. Specificially, we implement our approach for three typical NLP tasks: POS tagging (sequence-to-sequence learning, multi-class classification), chunking (sequence-to-sequence learning, multi-class classification), sentence sentiment analysis (sentence classification, binary classification).

{\bf Network Structure:} We use a very simple network structure. The neural network consists of an embedding layer, an ART layer as described in section~\ref{sec:lstm}, and a task-specific output layer for the prediction. We will elaborate the output layer and the loss function in each of the tasks below. We use 100d GloVe vectors~\citep{pennington2014glove} as the initialization for ART and all its ablations. %The target domain contains an embedding layer, an LSTM layer, and a output layer. The output layer varies in different applications.
%We compare our model with two baselines. Baseline1 is the model of open domain, we use it for close domain POS tagging. And baseline2 is similar to baseline1, but it is trained on close domain.

{\bf Competitor Models} We compare ART with the following ablations.
%We didn't compare with some very complicated neural networks (e.g. \cite{li2018hierarchical}) since we only use a simple ART neural network and want to verify the simple attentive recurrent transfer mechanism. We expect the performance improvement by applying ART to these complicated neural networks.
\begin{itemize}
\item {\bf LSTM} (no transfer learning): It uses a standard LSTM without transfer learning. It is only trained by the data of the target domain. %We use LSTM to verify the effectiveness of the ART.
\item {\bf LSTM-u} It uses a standard LSTM and is trained by the union data of the source domain and the target domain.
\item {\bf LSTM-s} It uses a standard LSTM and is trained only by the data of the source domain. Then parameters are used to predicting outputs of samples in the target domain.
%\item {\bf SDPA-NT}: is the SDPA without transfer. It consists of an embedding layer, a SDPA layer, and an output layer. It only trained by the target domain's data.
\item {\bf Layer-Wise Transfer (LWT)} (no cell-level information transfer): It consists of a layer-wise transfer learning neural network. More specifically, only the last cell of the RNN layer transfers information. This cell works as in ART. LWT only works for sentence classification. We use LWT to verify the effectiveness of the cell-level information transfer.
\item {\bf Corresponding Cell Transfer (CCT)} (no collocation information transfer): It only transfers information from the corresponding position of each cell. We use CCT to verify the effectiveness of collocating and transferring from the source domain. %consists of an aligned transfer layer of LSTM. That is, only the $i-1$-th cell in the source domain transfer the information to the $i$-th cell in the target domain.
%\item {\bf DANN}~\cite{ganin2016domain}: use adversarial training to learn the common representations and perform domain adaptation.
% it's output representation is a feature vector. And a further SVN is used for classification. Thus DANN cannot solve the sequence labeling task (i.e. POS tagging).
%\item {\bf Stanford POS tagger}~\cite{Toutanova:2003:FPT:1073445.1073478}: is one of the state-of-the-art model for POS tagging.
\end{itemize}

%Note that some competitor models only work for sentence classification (e.g.  DANN). So we only compare with them in the sentiment classification task.
We also compare ART with state-of-the-art transfer learning algorithms. For sequence labeling, we compare with hierarchical recurrent networks (HRN)~\citep{yang2017transfer} and FLORS~\citep{schnabel2014flors,wenpeng-2015-flors}. For sentence classification, we compare with DANN~\citep{ganin2016domain}, DAmSDA~\citep{ganin2016domain}, AMN~\citep{li2017end}, and HATN~\citep{li2018hierarchical}. Note that FLORS, DANN, DAmSDA, AMN and HATN use labeled samples of the source domain and unlabeled samples of both the source domain and the target domain for training. Instead, ART and HRN use labeled samples of both domains.

\subsection{Sentence Classification: Sentiment Analysis}

%Sentiment analysis solves the problem of identifying the sentiment of a sentence: {\it positive} or {\it negative}.
%We also evaluate the effectiveness of our approach in sentence classification. Specifically, we test it over the sentence level sentiment analysis problem.

{\bf Datasets:} We use the Amazon review dataset~\citep{blitzer2007biographies}, which has been widely used for cross-domain sentence classification. It contains reviews for four domains: {\it books, dvd, electronics, kitchen}. Each review is either positive or negative. We list the detailed statistics of the dataset in Table~\ref{tab:amazondata}. We use the training data and development data from both domains for training and validating. And we use the testing data of the target domain for testing.
%We use the Stanford Sentiment Treebank (SSTb)~\citep{socher2013recursive} as movie review corpus, and Stanford Twitter Sentiment corpus (STS)~\citep{go2009twitter} as twitter corpus. SSTb contains five sentiment tags: very negative, negative, neutral, positive and very positive. STS contains three sentiment tags: positive, negative and neutral. In this experiment, we only consider the binary sentiment analysis problem. For SSTb, we remove the neutral sentences, and merge the two negative classes and two positive classes into the negative class and positive class, respectively. In STS, we remove the neutral sentences and retain the rest two classes. The statistics of the datasets are shown in Table~\ref{tab:sentimentdata}.

\begin{table}[!htb]
\small
\vspace{-0.2cm}
\begin{center}
\caption{Statistics of the Amazon review dataset. {\it \%Neg.} denotes the ration of the negative samples. {\it Avg. L} denotes the average length of each review. {\it Vocab.} denotes the vocabulary size.}
\vspace{-0.2cm}
\begin{tabular}{  l  |  c  | c | c |  c  | c  |  c}
\hline
             &  Train &  Dev. & Test & $\%$ Neg. & Avg. L & Vocab. \\ \hline
\hline
Books          &  1400  &  200  &  400  & 50\%  &  159  & 62k            \\ \hline
Electronics          &  1398  &  200  &  400  & 50\% & 101  & 30k             \\ \hline
DVD            &  1400  &  200  &  400  & 50\% & 173  & 69k     \\ \hline
Kitchen        &  1400  &  200  &  400  & 50\% & 89  &  28k      \\ \hline
%Video          &  1400  &  200  &  400  & 50\% & 156  & 57k      \\ \hline
\end{tabular}
\vspace{-0.3cm}
\label{tab:amazondata}
\end{center}
%\vspace{-0.5cm}
\end{table}

{\bf Model Details:} To adapt ART to sentence classification, we use a max pooling layer to merge the states of different words. Then we use a perception and a sigmoid function to score the probability of the given sentence being positive. We use binary cross entropy as the loss function. The dimension of each LSTM is set to 100. We use the Adam~\citep{kingma2014adam} optimizer. We use a dropout probability of 0.5 on the max pooling layer.

\nop{
 sigmoid function for the hidden state of the last cell in the recurrent layer. It scores the probability of the given sentence $x$ being positive:
\begin{equation}
\begin{aligned}
\hat{y}^l = \sigma(W_y^l h_n^l + b_y^l)
\end{aligned}
\end{equation}
where $h_n^l$ is the hidden state of the last cell in the RNN layer, $l \in \{S,T\}$ indicates the domain, $\sigma$ is the sigmoid function, $W_y^l, b_y^l$ are the parameters of the output layers. We use binary cross entropy as the loss function.
}

\begin{table*}[!htb]
\small
\vspace{-0.2cm}
\begin{center}
\caption{Classification accuracy on the Amazon review dataset. }%The results of DANN is from~\cite{ganin2016domain} and~\cite{li2018hierarchical} (for Video).
\vspace{-0.2cm}
\tabcolsep=0.11cm
\begin{tabular}{  c  c   |  c  c  c  c   c   c c   c  c  c  }
\hline
    Source &   Target    &  LSTM & LSTM-u & LSTM-s & CCT & LWT & DANN & DAmSDA & AMN & HATN & ART  \\ \hline
\hline
Books &  DVD           & 0.695 & 0.770 & 0.718 & 0.730 & 0.784 & 0.725 & 0.755 & 0.818 & 0.813 & {\bf 0.870}         \\
Books &  Elec.   & 0.733 & 0.805 & 0.678 & 0.768 & 0.763 & 0.690 & 0.760 & 0.820 & 0.790 & {\bf 0.848}          \\
Books &  Kitchen       & 0.798 & 0.845 & 0.678 & 0.818 & 0.790 & 0.770 & 0.760 & 0.810 & 0.738 & {\bf 0.863}     \\ \hline
%Books &  Video         & 0.845 & 0.848 & 0.815 & 0.832& {\bf 0.868}         \\ \hline
DVD &  Books           & 0.745 & 0.788 & 0.730 & 0.800 & 0.778 & 0.745 & 0.775 & 0.825 & 0.798 & {\bf 0.855}           \\
DVD &  Elec.     & 0.733 & 0.788 & 0.663 & 0.775 & 0.785 & 0.745 & 0.800 & 0.810 & 0.805 & {\bf 0.845}       \\
DVD &  Kitchen         & 0.798 & 0.823 & 0.708 & 0.815 & 0.785 & 0.780 & 0.775 & 0.830 & 0.765 & {\bf 0.853}       \\ \hline
%DVD &  Video           & 0.845 & 0.855 & 0.835 & 0.860& {\bf 0.868}        \\ \hline
Elec. &  Books   & 0.745 & 0.740 & 0.648 & 0.773 & 0.735 & 0.655 & 0.725 & 0.785 & 0.763 & {\bf 0.868}        \\
Elec. &  DVD     & 0.695 & 0.753 & 0.648 & 0.768 & 0.723 & 0.720 & 0.695 & 0.780 & 0.788 & {\bf  0.855}        \\
Elec. &  Kitchen & 0.798 & 0.863 & 0.785 & 0.823 & 0.793 & 0.823 & 0.838 & {\bf 0.893} & 0.808 & 0.890       \\ \hline
%Electronics &  Video   & 0.845 & 0.835 & 0.793 & 0.772 & {\bf 0.853}       \\ \hline
Kitchen &  Books       & 0.745 & 0.760 & 0.653 & 0.803 & 0.755 & 0.645 & 0.755 & 0.798 & 0.740 & {\bf 0.845}          \\
Kitchen &  DVD         & 0.695 & 0.758 & 0.678 & 0.750 & 0.748 & 0.715 & 0.775 & 0.805 & 0.738 & {\bf 0.858}         \\
Kitchen &  Elec. & 0.733 & 0.815 & 0.758 & 0.810 & 0.805 & 0.810 & {\bf 0.870} & 0.833 & 0.850 & 0.853         \\ \hline
%Kitchen &  Video       & 0.845 & 0.835 & 0.818 & 0.764& {\bf 0.853}         \\ \hline
%Video &  Books         & 0.745 & 0.800 & 0.790 & 0.800& {\bf 0.855}        \\
%Video &  DVD           & 0.695 & 0.785 & 0.753 & 0.842& {\bf 0.858}         \\
%Video &  Electronics   & 0.733 & 0.813 & 0.775 & 0.757& {\bf 0.830}         \\
%Video &  Kitchen       & 0.798 & 0.825 & 0.745 & 0.752& {\bf 0.865}         \\ \hline
\multicolumn{2}{c|}{ Average} & 0.763 & 0.792 & 0.695 & 0.803 & 0.774 & 0.735 & 0.774 & 0.817 & 0.783 & {\bf 0.858}      \\ \hline
\end{tabular}
%\vspace{-0.2cm}
\label{tab:amazonresult}
\end{center}
\vspace{-0.3cm}
\end{table*}

{\bf Results:} We report the classification accuracy of different models in Table~\ref{tab:amazonresult}. The no-transfer LSTM only performs accuracy of $76.3\%$ on average. ART outperforms it by $9.5\%$. ART also outperforms its ablations and other competitors. This overall verifies the effectiveness of ART.
%Since we use a very simple network structure with ART, more improvement is expected by applying it to more complicated neural networks. We also evaluate the effectiveness of some detailed ART designs below.

{\bf Effectiveness of Cell-Level Transfer} LWT only transfers layer-wise information and performs accuracy of $77.4\%$ on average. But ART and CCT transfer more fine-grained cell-level information. CCT outperforms LWT by $2.9\%$. ART outperforms LWT by $8.4\%$. This verifies the effectiveness of the cell-level transfer.

%The only difference between ART and LW-TL is, the former one transfer information for all recurrent cells. LW-TL still gains from the layer-wise transfer that it outperforms LSTM-NT with $1.1\%$ on average. But by using cell-level transfer, ART-LSTM outperforms LW-TL by $4.8\%$ on average. So even for sentence classification, cell-level transfer gains from the fine-grained cell-level information transfer. This verify the effectiveness of the cell-level transfer in ART.

{\bf Effectiveness of Collocation and Transfer} CCT only transfers the corresponding position's information from the source domain. It achieves accuracy of $80.3\%$ on average. ART outperforms CCT by $5.5\%$ on average. ART provides a more flexible way to transfer a set of positions in the source domain and represent the long-term dependency. This verifies the effectiveness of ART in representing long-term dependency by learning to collocate and transfer.

%This is used to some the long-dependency limitation in RNN. To verify the effectiveness of the attentive transfer mechanism, we compare ART-LSTM with AT-TL. Note AT-TL only transfer the aligned cell's information. Such a transfer mechanism ($\%$) still works compared with the no-transfer LSTM($76.3\%$). But by attentively transfer all source cells' information, ART-LSTM outperforms AT-TL by $\%$ on average. So the attentive transfer successfully transfer information from a set of positions in the source domain. This verifies the effectiveness of the attentive transfer in ART.

%{\bf Generality of ART} We apply ART to different neural networks, including ART-LSTM and cART-SDPA. From the results, ART-LSTM outperforms LSTM-NT by $5.9\%$. And cART-SDPA outperforms SDPA-NT by $\%$. Here LSTM is an RNN architecture. And SDPA is an ``attention-only'' architecture. The success in different types of network architectures verifies that ART can be generalized to different architectures. We owe this to the simple mechanism of ART, which makes it flexible for different and complicated neural networks.

{\bf Minimally Supervised Domain Adaptation} {\cwy  We also evaluate ART when the number of training samples for the target domain is much fewer than that of the source domain. For each target domain in the Amazon review dataset, we combine the training/development data of rest three domains as the source domain. We show the results in Table~\ref{tab:amazonminimally}. ART outperforms all the competitors by a large margin. This verifies its effectiveness in the setting of minimally supervised domain adaptation.}

\begin{table*}[!htb]
\small
\vspace{-0.2cm}
\begin{center}
\caption{Classification accuracy with scarce training samples of the target domain.}
\vspace{-0.2cm}
\begin{tabular}{  l  |  c c  c c  c c c c  c}
\hline
Target  & LSTM & LSTM-u & LSTM-s & CCT & LWT & HATN & ART \\ \hline
\hline
Books & 0.745 & 0.813 &  0.800 & 0.848 & 0.808 & 0.820 & {\bf 0.895} \\
DVD & 0.695 & 0.748 & 0.795 & 0.870 & 0.770 & 0.828 & {\bf 0.875} \\
Electronics & 0.733 & 0.823 & 0.760 & 0.848 & 0.818 & 0.863 & {\bf 0.865} \\
Kitchen & 0.798 & 0.840 & 0.795 & 0.860 & 0.840 & 0.833 & {\bf 0.870} \\ \hline
Average & 0.743 & 0.806 & 0.788 & 0.856 & 0.809 & 0.836 & {\bf 0.876} \\
\hline
\end{tabular}
\label{tab:amazonminimally}
\end{center}
\vspace{-0.3cm}
\end{table*}

\subsection{Sequence Labeling}
%The goal of POS tagging is to assign part-of-speech tags to each word of the given sentence. We test our approach over the People's Daily corpus, which is a Chinese POS tagging corpus with 294240 sentences and 44 tags.

We evaluate the effectiveness of ART w.r.t. sequence labeling. We use two typical tasks: POS tagging and NER (named entity recognition). The goal of POS tagging is to assign part-of-speech tags to each word of the given sentence. The goal of NER is to extract and classify the named entity in the given sentence.

{\bf Model Settings:} POS tagging and NER are multi-class labeling tasks. To adapt ART to them, we follow HRN~\citep{yang2017transfer} and use a \zgy{CRF layer} to compute the tag distribution of each word. We predict the tag with maximized probability for each word.
\nop{
Specifically, the output layer predicts the tag by:
\begin{equation}
\begin{aligned}
&\hat{p}_\theta^l(y_i) = softmax(W^l h^l_i + b^l),\\
&\hat{y}_i^l = \mathop{\argmax}_{y}\hat{p}_\theta^l(y)
\end{aligned}
\end{equation}
where $\hat{y}_i^T$ is the target domain's output, $W_T, b^T$ are the parameters of the layers, $l \in \{S,T\}$ is the domain indicator.
}
We use categorical cross entropy as the loss function. The dimension of each LSTM cell is set to 300. By following HRN, we use \zgy{the concatenation of} 50d word embeddings and 50d character embeddings as the input of the ART layer. \zgy{We use 50 1d filters for CNN char embedding, each with a width of 3.} We use the Adagrad~\citep{duchi2011adaptive} optimizer. We use a dropout probability of 0.5 on the max pooling layer.

{\bf Dataset:} We use the dataset settings as in~\cite{yang2017transfer}. For POS Tagging, we use Penn Treebank (PTB) POS tagging,
%Genia biomedical corpus~\citep{kim2003genia},
and a Twitter corpus~\citep{ritter2011named} as different domains. For NER, we use CoNLL 2003~\citep{tjong2003introduction} and Twitter~\citep{ritter2011named} as different domains. Their statistics are shown in Table~\ref{tab:sequencedata}. By following~\cite{ritter2011named}, we use 10\% training samples of Twitter (Twitter/0.1),
1\% training samples of Twitter (Twitter/0.01),
%0.1\% of Genia (Genia/0.001),
and 1\% training samples  of CoNLL (CoNLL/0.01) as the training data for the target domains to simulate a low-resource setting. Note that the label space of these tasks are different. So some baselines (e.g. LSTM-u, LSTM-s) cannot be applied.

\setlength{\tabcolsep}{2pt}
\begin{table}[!htb]
\small
\vspace{-0.2cm}
\begin{center}
\caption{Dataset statistics.} %and results of POS tagging. ``B.1'' stands for baseline 1. ``B.2'' stands for baseline 2. ``St.'' stands for Stanford POS tagger. ``T-$L_c$'' stands for the concatenation version of the T-LSTM. ``T-$L_g$'' stands for the gated version of the T-LSTM.
\vspace{-0.2cm}
\begin{tabular}{  l  |  c  | c c  c }
\hline
Benchmark     & Task  & \# Training Tokens     & \# Dev Tokens &  \# Test Tokens \\ \hline
\hline
PTB    &  POS Tagging   &   912,344    & 131,768        & 129,654 \\
%Genia  &  POS Tagging   &   400,658    & 50,525        & 49,761 \\
Twitter  &  POS Tagging   &   12,196    & 1,362        & 1,627 \\ \hline
CoNLL 2003 & NER       &    204,567  &  51,578  &  46,666 \\
Twitter  &  NER      &    36,936  &  4,612  &  4,921  \\ \hline
\end{tabular}
\label{tab:sequencedata}
\end{center}
\vspace{-0.2cm}
\end{table}

\begin{table}[!htb]
\small
\vspace{-0.3cm}
\begin{center}
\caption{Performance over POS tagging and NER.} %and results of POS tagging. ``B.1'' stands for baseline 1. ``B.2'' stands for baseline 2. ``St.'' stands for Stanford POS tagger. ``T-$L_c$'' stands for the concatenation version of the T-LSTM. ``T-$L_g$'' stands for the gated version of the T-LSTM.}
\vspace{-0.2cm}
\begin{tabular}{  l  |  c   c |c  c c c c }
\hline
Task & Source     & Target  & HRN & FLORS & LSTM & CCT & ART \\ \hline
\hline
POS Tagging   &  PTB    &  Twitter/0.1   &   0.837  &  \multirow{2}{*}{0.763}  &  0.798  &  0.852    & {\bf 0.859} \\
POS Tagging   &  PTB    &  Twitter/0.01   &  0.647 &    & 0.573  & 0.653     &   {\bf 0.658} \\ \hline
%POS Tagging    &  PTB  &  Genia/0.01   & 0.963     &    &     & \\
%POS Tagging    &  PTB  &  Genia/0.1   &  0.982    &    &     &  \\ \hline
NER           &  CoNLL & Twitter/0.1   &   0.432  &  - &  0.210    &  0.434  & {\bf 0.450} \\
NER           &  Twitter & CoNLL/0.01  &  - &  - & 0.576          & 0.675  &  {\bf 0.707}  \\ \hline
\end{tabular}
\label{tab:postag}
\end{center}
\vspace{-0.2cm}
\end{table}

Table~\ref{tab:postag} shows the per-word accuracy (for POS tagging) and f1 (for NER) of different models. From the table, we see that the performances of all tasks are improved by ART. For example, when transferring from PTB to Twitter/0.1, ART outperforms HRN by 2.2\%. ART performs the best among all competitors in almost all cases. This verifies the effectiveness of ART w.r.t. sequence labeling. Note that FLORS is independent of the target domain. If the training corpus of the target domain is quite rare (Twitter/0.01), FLORS performs better. But with richer training data of the target domain (Twitter/0.1), ART outperforms FLORS. %Similar to the results of sentiment analysis, the results also verifies the effectiveness of cell-level transfer and the effectiveness of collocation.

\vspace{-0.2cm}

\subsection{Visualization of the Aligned Transfer}

\begin{figure}[ht]
\centering
    \begin{subfigure}[b]{0.46\textwidth}
    \hspace{-1cm}
        \centering
            \includegraphics[scale=0.45]{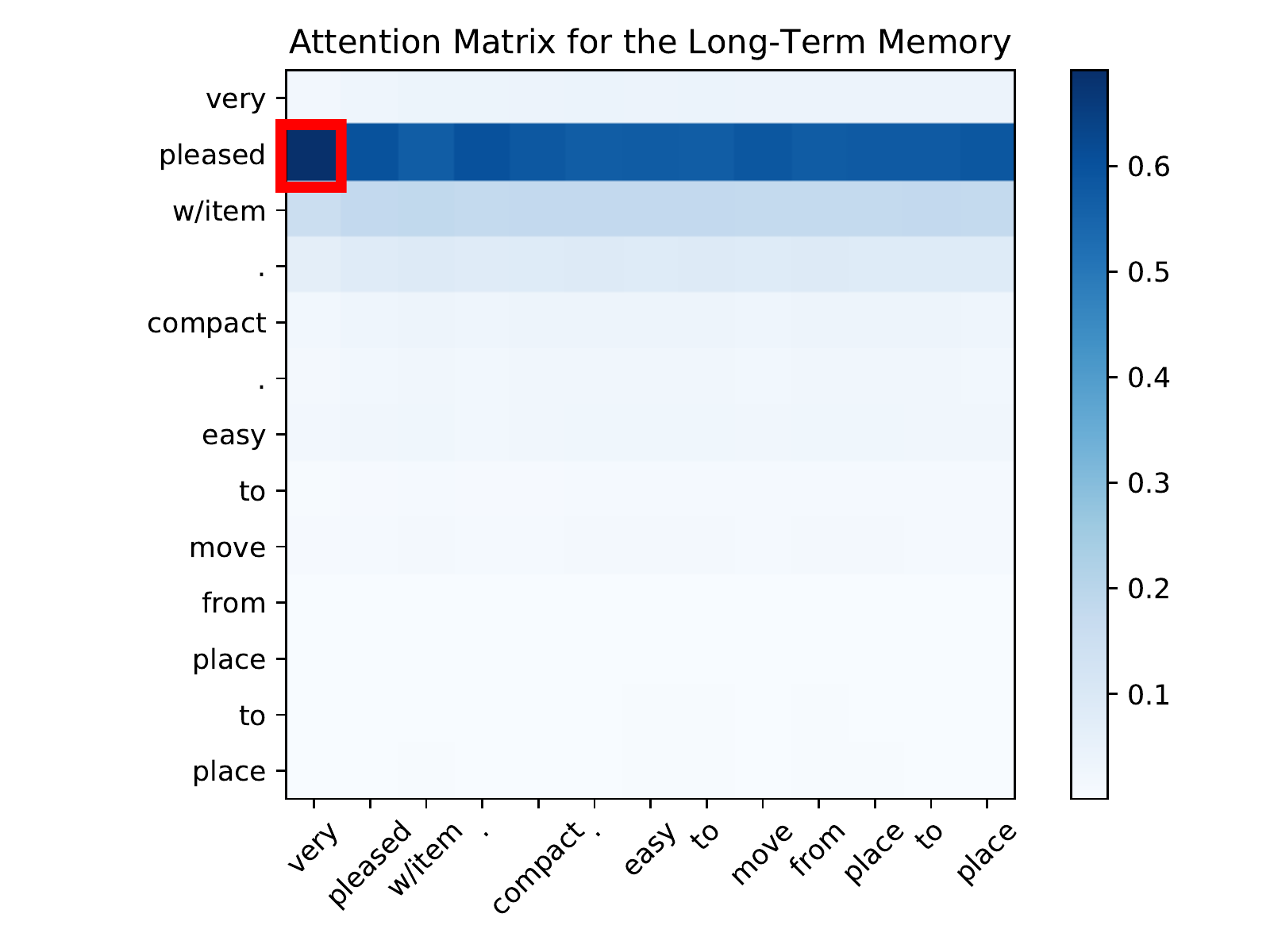}
            %\caption{Attention matrix for the long-term memory.}
            %$O_1$ is layer for the open domain. $S_1$ and $S_2$ are layers for the specific domain. The dash line means ``kitt'' is trained insufficiently.
        \label{fig:attention:a}
    \end{subfigure}
%    \newline
    \begin{subfigure}[b]{0.46\textwidth}
        \centering
            \includegraphics[scale=0.45]{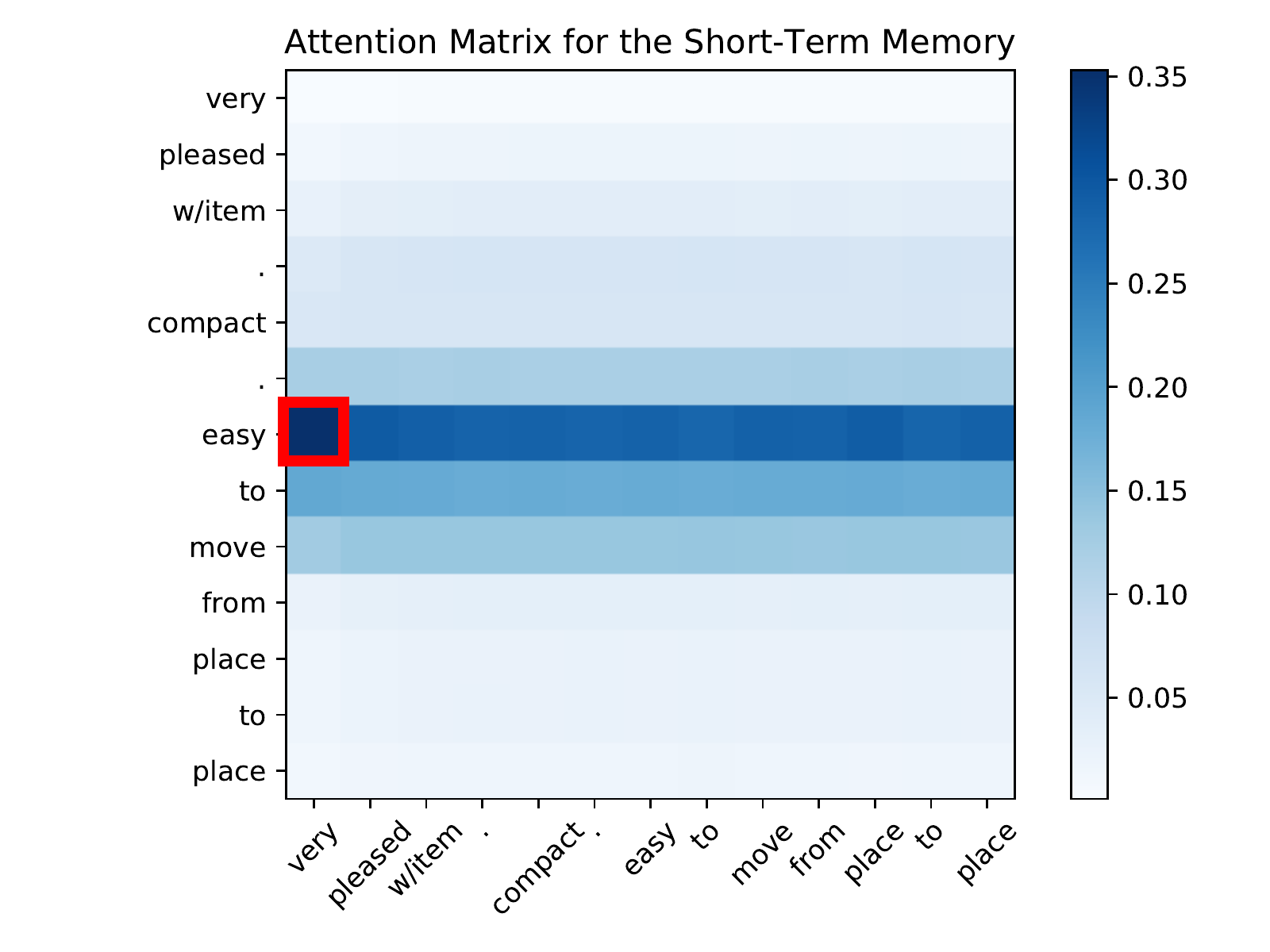}
            %\caption{Attention matrix for the short-term memory.}
        \label{fig:attention:b}
    \end{subfigure}
%    \begin{subfigure}[b]{.45\linewidth}
%        \centering
%            \includegraphics[scale=0.6]{figs/transferablernn.pdf}
%            \caption{The unit-by-unit mechanism for sequence-to-sequence learning.}
%        \label{fig:modelillustration:unitbyunit}
%    \end{subfigure}
\vspace{-0.7cm}
\caption{Attention matrix visualization. The x-axis and the y-axis denote positions in the target domain and source domain, respectively. Figure (a) shows the attention matrix for the long-term memory in the forward neural network. Figure (b) shows the attention matrix for the short-term memory in the forward neural network.}
\label{fig:attentionmatrix} %% label for entire figure
\vspace{-0.5cm}
\end{figure}

ART aligns and transfers information from different positions in the source domain. Intuitively, we use the alignment and attention matrices to represent cross-domain word dependencies. So positions with stronger dependencies will be highlighted during the transfer. We visualize the attention matrix for sentiment analysis to verify this. We show the attention matrices for the short-term memory $h$ and for the long-term memory $c$ in figure~\ref{fig:attentionmatrix}.

{\bf Effectiveness of the Cell-Level Transfer} From figure~\ref{fig:attentionmatrix}, words with stronger emotions have more attentions. For example, the word ``pleased'' for the long-term memory and ``easy to move'' for the short-term memory have strong attention for almost all words, which sits well with the intuition of cell-level transfer. The target domain surely wants to accept more information from the meaningful words in the source domain, not from the whole sentence. Notice that $c$ and $h$ have different attentions. Thus the two attention matrices represent discriminative features.

{\bf Effectiveness in Representing the Long-Term Dependency} We found that the attention reflects the long-term dependency. For example, in figure~\ref{fig:attentionmatrix}~(b), although all words in the target domain are affected by the word ``easy'', the word ``very'' has highest attention. This makes sense because ``very'' is actually the adverb for ``easy'', although they are not adjacent. ART highlights cross-domain word dependencies and therefore gives a more precise understanding of each word.
\vspace{-0.2cm}

\vspace{-0.15cm}
\section{Related work}
\vspace{-0.15cm}

{\bf Neural network-based transfer learning} {\cwy The layer-wise transfer learning approaches~\citep{glorot2011domain,ajakan2014domain,DBLP:conf/acl/ZhouXHH16} represent the input sequence by a non-sequential vector. These approaches cannot be applied to seq2seq or sequence labeling tasks. To tackle this problem, algorithms must transfer cell-level information in the neural network~\citep{yang2017transfer}. Some approaches use RNN to represent cell-level information. \citet{ying2017recurrent} trains the RNN layer by domain-independent auxiliary labels. \citet{ziser2018pivot} trains the RNN layer with pivots. However, the semantics of a word can depend on its collocated words. These approaches cannot represent the collocated words. In contrast, ART successfully represents the collocations by attention.}

{\bf Pre-trained models} {\cwy ART uses a pre-trained model from the source domain, and fine-tunes the model with additional layers for the target domain. Recently, pre-trained models with additional layers are shown to be effectiveness for many downstream models (e.g. BERT~\citep{devlin2018bert}, ELMo~\citep{peters2018deep}). As a pre-trained model, ELMo uses bidirectional LSTMs to generate contextual features. Instead, ART uses attention mechanism in RNN that each cell in the target domain directly access information of all cells in the source domain. ART and these pre-trained models have different goals. ART aims at transfer learning for one task in different domains, while BERT and ELMo focus on learning general word representations or sentence representations. } %The architecture of ART is much simpler than BERT or ELMo. So it does not require very large scale training data for the pre-trained model. In our experiments, ART works well when the source domain only has 1400 training samples.  %And besides using
 %These approaches use Transformer~\citep{vaswani2017attention} to represent the sequence, while ART uses RNN. And ART modifies RNN

%We decide to use DNN for the transfer learning, due to its impressive performance in recent NLP applications. \citep{collobert2011natural} uses a convolution neural network for sentence modeling. The work performs well in many NLP tasks, including POS tagging, chunking, named entity recognition and semantic role labeling. Variances of DNN are also used in different applications and exhibit great values. For example, \cite{DBLP:journals/corr/LiuQH16a} uses coupled-LSTM for matching question against answer. \cite{luong2013better} uses recursive neural networks for word representation learning. All these advantages inspire us to use DNN for transfer learning in NLP applications.
\vspace{-0.15cm}

\section{Conclusion}
\vspace{-0.15cm}
In this paper, we study the problem of transfer learning for sequences. We proposed the ART model to collocate and transfer cell-level information. ART has three advantages: (1) it transfers more fine-grained cell-level information, and thus can be adapted to seq2seq or sequence labeling tasks; (2) it aligns and transfers a set of collocated words in the source sentence to represent the cross domain long-term dependency; (3) it is general and can be applied to different tasks. Besides, ART verified the effectiveness of pre-training models with the limited but relevant training corpus. %We conducted extensive experiments over sentence classification (sentiment analysis) and sequence labeling (POS tagging and NER). ART outperforms all competitors over all tasks. %Our approach outperforms other competitors.

%In this paper, we study the problem of transfer learning in NLP applications with. We proposed an LSTM network based on PNN framework. Our approach uses two neural networks for open domain modeling and specific domain modeling, respectively. These neural networks share two layers. The specific domain's neural network uses the transferred knowledge from the overlapping layers. The knowledge are transferred through lateral connections from the overlapping layers. We gave the detailed implementation of our approach. And we conducted extensive experiments over typical NLP tasks, including sentiment analysis, POS tagging, and chunking. Our approach outperforms other competitors. The results of all these experiments show our approach really transfers open domain knowledge to improve the results, and can be adapted to different NLP tasks.

%We think there are still many potential values for the model, and many other applications to be applied. Note that the PNN framework uses a very simple way (lateral connection) for knowledge transfer. So basically any multi-layer DNN framework can be used over this framework. We expect to use bidirectional LSTM model or attention LSTM model to get better results. This will be our future work. And we expect to employ our approach in more NLP applications.

\subsubsection*{Acknowledgments}
Guangyu Zheng and Wei Wang were supported by Shanghai Software and Integrated Circuit Industry Development Project(170512).

\bibliographystyle{iclr2019_conference}
\bibliography{paper}

\end{document}